\documentclass[10pt,twocolumn,letterpaper]{article}

\usepackage{iccv}
\usepackage{times}
\usepackage{epsfig}
\usepackage{graphicx}
\usepackage{amsmath}
\usepackage{amssymb}
\usepackage{kotex}
\usepackage[normalem]{ulem}
\usepackage{multirow}
\usepackage{array}
\usepackage{bm}

\usepackage[breaklinks=true,bookmarks=false]{hyperref}

\iccvfinalcopy 

\def\iccvPaperID{****} 

\ificcvfinal\fi

\begin{document}

\title{Utilizing Task-Generic Motion Prior \\ to Recover Full-Body Motion from Very Sparse Signals}

\author{Myungjin Shin \\
Yonsei University \\
{\tt\small mjsh25@yonsei.ac.kr}
\and
Dohae Lee\\
Yonsei University\\
{\tt\small dlehgo1414@yonsei.ac.kr}
\and
In-Kwon Lee\\
Yonsei University\\
{\tt\small iklee@yonsei.ac.kr}}

\maketitle
\ificcvfinal\thispagestyle{empty}\fi

\begin{abstract}

The most popular type of devices used to track a user's posture in a virtual reality experience consists of a head-mounted display and two controllers held in both hands. However, due to the limited number of tracking sensors (three in total), 
faithfully recovering the user in full-body is challenging, limiting the potential for interactions among simulated user avatars within the virtual world.
Therefore, recent studies have attempted to reconstruct full-body poses using neural networks that utilize previously learned human poses or accept a series of past poses over a short period.
In this paper, we propose a method that utilizes information from a neural motion prior to improve the accuracy of reconstructed user's motions. Our approach aims to reconstruct user's full-body poses by predicting the latent representation of the user's overall motion from limited input signals and integrating this information with tracking sensor inputs. This is based on the premise that the ultimate goal of pose reconstruction is to reconstruct the motion, which is a series of poses. Our results show that this integration enables more accurate reconstruction of the user's full-body motion, particularly enhancing the robustness of lower body motion reconstruction from impoverished signals. 
Web: \texttt{https://https://mjsh34.github.io/mp-sspe/}.
\end{abstract}


\section{Introduction}
    The technology of today's Mixed Reality (MR) has extended traditional interpersonal experiences into the virtual realm, from social gatherings and gaming, to collaborative work, just to name a few. While these experiences have traditionally been limited to settings in which all participants are present in the same environment, MR systems instead rely on virtual avatars to simulate the experience and benefits of non-verbal communication. Unfortunately, real-time data streams provided by commercial MR devices, which typically consist of a head-mounted display (HMD) and two hand-held controllers, each tracked in a small room-scale grid, are insufficient to accurately reproduce the full pose of its user, often resulting in MR environments where avatars only show head and hands. Studies have shown that a hand-only avatar provide little sense of embodiment \cite{8263407, 10.5555/3061323.3061345, 8998305}, whereas a full-body avatar can significantly enhance the user experience by creating better sense of embodiment and presence \cite{10.5555/3061323.3061345, 8998305, heidickerpaul}.

    Tasked with the challenge of recovering full-body posture of humans from sparse signal streams, previous works have attempted to reconstruct the human body from tracking signals from four or more joints including the pelvis \cite{DIP:SIGGRAPHAsia:2018, 10.1111/cgf.13131, amass_MoSh, lobstr, pip, TransPoseSIGGRAPH2021}, and from ego-centric cameras \cite{Jiang_2017_CVPR, Yuan_2019_ICCV, Ng_2020_CVPR, Yuan_2021_CVPR}, which are unavailable on most MR devices at the present day. Recent works have attempted to reconstruct full-body poses from pose information alone from the HMD and handheld controllers \cite{coolmoves, vaehmd, flag, avatarposer, cmm, questsim, n3p}. However when the the reconstructed poses are combined to form a complete motion, they often lead to unnatural motions that fail to match the user's desired action. These systems also often fail to faithfully reproduce lower body motion beyond basic actions such as standing still, and walking at various speeds. 

    \begin{figure}
    \centering
    \includegraphics[width=8.25cm]{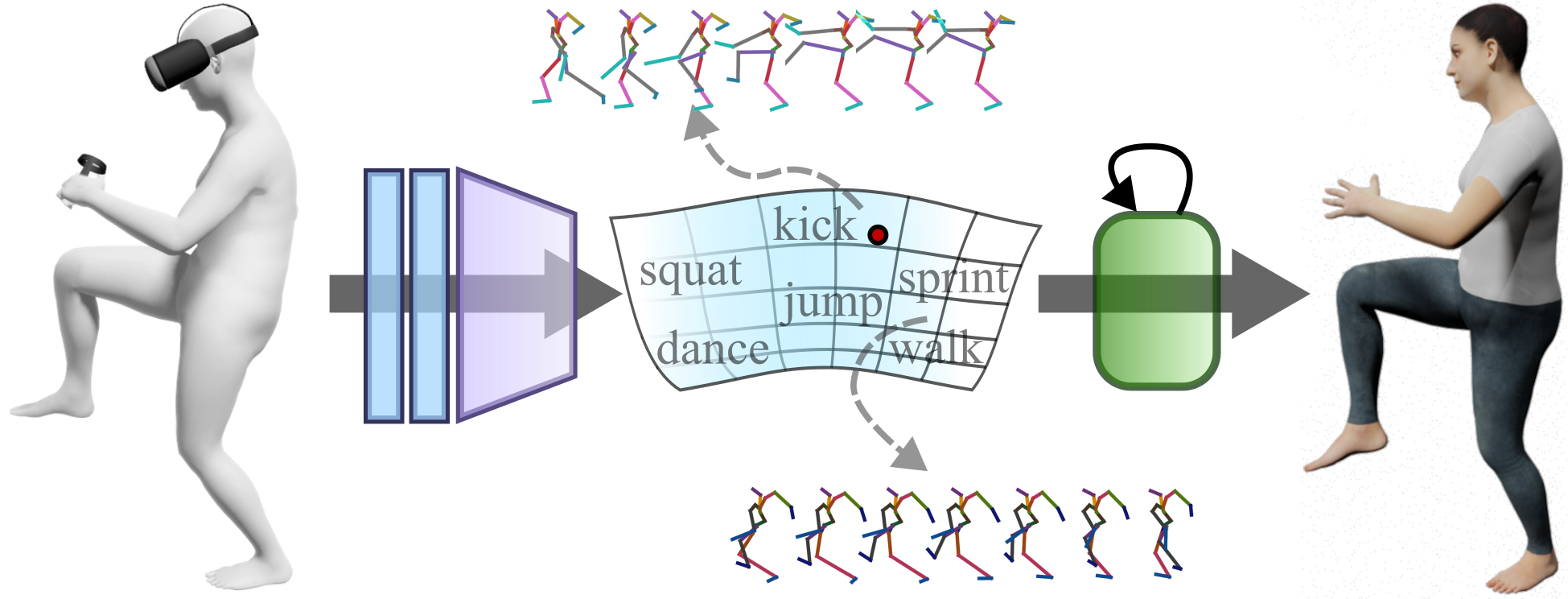}
    \caption{We present a method that utilizes a motion prior to encode the overall motion of a user for full-body pose reconstruction, using only the information of head and two hands.
    }
    \label{fig:intro}
    \end{figure}

    We propose a method effectively utilizing a task-generic neural motion prior \cite{hmvae, nemf, humor, motionclip} (Section \ref{tgmp}) aimed at solving the issues mentioned above. We exploit a generative motion prior model with an encoder-decoder architecture that is initially trained to reconstruct full-body motion while learning a latent space of human motions. We train a motion encoder to predict latent representations of motion from a sequence of sparse poses obtained from the three sources mentioned earlier, utilizing latent space learned by motion prior. Finally, we train a sequence (time-series) model that generates full-body pose from the sparse pose sequence and the latent representations of the overall motion. We achieve the following:

    \begin{itemize}
        \item Our method utilizing a motion prior outperforms state-of-the-art methods in reconstructing a full-body pose at a single frame, and in reconstructing motions from combined full-body pose reconstructions from three tracking signals. We evaluate static pose reconstruction and full motion reconstruction performances using appropriate metrics. 
        \item We show that our method produces natural-looking motions which match the intended action of the underlying full-body motion.
        \item Our method improves on reconstructing lower body motions which methods without any prior on motion struggle at.
    \end{itemize}

    We show our model's superior performance against previous works using a diverse set of quantitative metrics (Table \ref{table:main_results}), user studies (Tables \ref{table:survey1_results_pref} and \ref{table:survey1_results_mm}, and Figure \ref{fig:survey2_results}), and qualitative evaluation (Figure \ref{fig:qual_eval_2anims}).

\section{Related Works}
\subsection{Human Body Representations}
The SMPL model \cite{smpl} represents the human body as a kinematic tree consisting of 24 joints using two parameters: $\boldsymbol{\theta}$ and $\boldsymbol{\beta}$, where $\boldsymbol{\theta} \in \mathbb{R}^{24 \times 3}$ represents the rotations of all 24 joints in the axis-angle representation, and $\boldsymbol{\beta} \in \mathbb{R}^{10}$ represents the shape parameter that describes the body type derived via principle component analysis \cite{smpl} for each gender (male, female, and neutral). We parametrize the full-body pose using the joint rotations in 6D form, which is a continuous representation of 3D rotation proposed by Yi et al.~\cite{6drot} to be effective in training neural networks (also used by previous works on the same task \cite{vaehmd, flag, avatarposer}).
For training and inference, we use the neutral body with mean body shape: $\boldsymbol{\beta} = \boldsymbol{0}$, disregarding variations in body shape, similar to the approach taken in previous works \cite{vaehmd, flag, avatarposer} which also do not consider body shape diversity.

\subsection{Full-Body Reconstruction from Various Signal Streams} \label{fbr_from_ss}
An abundance of literature is dedicated to the recovery of full-body pose from observations of various modalities such as images \cite{Bogo:ECCV:2016, omran2018nbf, Zanfir_2021_ICCV, pymafx2022}, videos \cite{VNect_SIGGRAPH2017, shi2020motionet, Yu:2021:MovingCam, Dabral_2021_ICCV, iskakov2019learnable}, and sparsely-worn body trackers \cite{10.1111/cgf.13131, DIP:SIGGRAPHAsia:2018, TransPoseSIGGRAPH2021, lobstr, pip}.
Notably, the last set of research, while similar to our problem setting, work with richer information by the tracking the pelvis at the very least and often the lower body as well \cite{10.1111/cgf.13131, DIP:SIGGRAPHAsia:2018, TransPoseSIGGRAPH2021,  pip}.

\subsection{Full-Body Reconstruction from Head and Hands}
Unlike most previous works that focus on recovering full-body pose from sparse body-worn trackers introduced in Section \ref{fbr_from_ss}, efforts have been made to make use of only three tracking signals, namely the positions and rotations of the head and hands, which most commercial MR devices provide \cite{coolmoves, vaehmd, flag, avatarposer, cmm, questsim, n3p}.

We categorize the recent lines of works into four types: 
(1) Motion Matching: Given sparse observations, these works \cite{coolmoves, cmm} attempt to find the most fitting motion from a predefined animation database. Their primary goals lie not in not precise reconstruction of the full-body, but in accentuating and stylizing motion.
(2) Physics-Based Simulation: Recently, QuestSim \cite{questsim} and Neural3Points \cite{n3p} have been proposed for simulating full-body avatars by predicting parameters of physical simulation, rendering motions based on laws of physics. 
In their studies \cite{questsim, n3p}, the authors observed that while the synthesized motions are physically plausible, they can exhibit stiffness and unnaturalness. These models also encounter challenges when attempting to replicate complex lower-body motions that have low correlation with their corresponding upper-body movements.
Moreover, they are susceptible to deviating from actual movements as errors in physical simulation accumulate and falling over, in which cases simulation needs to be restarted.
Neural3Points \cite{n3p} attempts to mitigate the last issues by using a direct full-pose prediction model in conjunction, but still resorts to restarting the simulation if too much errors accumulate, compromising the realism and accuracy of generated motions.
The remaining two lines of work focus on directly predicting full-body pose at every frame.
(3) Sequence (Time-Series) Model for Full-Body Pose Estimation: AvatarPoser \cite{avatarposer} proposes using a Transformer Encoder \cite{transformer} to parse a 40-frame sequence of sparse pose signals to predict the full-pose.
(4) Generative Latent Space-Based Full-Body Pose Estimation: VAE-HMD \cite{vaehmd} and FLAG \cite{flag} rely on a decoder of a pose prior to predict the full-pose given a latent code derived from sparse pose signals at the current frame or over a short sequence of past frames. For full-pose priors VAE-HMD uses $\beta$-VAE \cite{betavae} for encoder and decoder and FLAG uses RealNVP \cite{realnvp} for encoder and decoder. To estimate latent codes from sparse signals (as substitute for full-pose encoders) VAE-HMD optimizes a $\beta$-VAE objective with a new encoder and FLAG employs a Transformer-based \cite{transformer} predictor.

We have found (3) sequence model-based methods to produce smoother motions than (4) generative latent space-based methods, while the latter tend to more accurately depict full-body pose at a given frame.
Our method integrates (3) and (4)'s approaches, predicting full-body pose using a sequence model while simultaneously making use of a generative latent space of motion to produce smooth and accurate full-body motions. We also utilize an explicit motion prior as opposed to static pose priors used by \cite{vaehmd, flag}.

\subsection{Task-Generic Motion Priors} \label{tgmp}
The term task-generic motion prior was first used in HM-VAE \cite{hmvae}, to describe ``a generalized motion prior, that learns complex human body motions from high-fidelity motion capture data". Unlike task-specific motion priors which are optimized for a single task, such as motion recovery from videos \cite{tcmr, vibe, meva}, task-generic motion priors \cite{nemf, hmvae, humor, motionclip} are generative models that can perform an array of motion-related tasks, such as motion interpolation (in-betweening), completion, synthesis, refinement, and even recovery of full-pose from partial observations via test-time optimization. NeMF \cite{nemf} categorizes these motion priors into two categories: time-series models and space-time models. Time-series models predict future motions based on past observations and are typically autoregressive, with HuMoR \cite{humor} being an example. Space-Time models, on the other hand, directly model the spatio-temporal kinematic state, often by taking in a whole motion as input at once \cite{nemf, hmvae, motionclip}.

We select MotionCLIP \cite{motionclip} for our full-body motion prior, a space-time motion prior with an auto-encoder \cite{goodfellow2016deep} architecture. MotionCLIP learns to embed the input motion in the latent space of CLIP \cite{openaiclip}, a large-scale neural network trained jointly on image and text. The CLIP space has demonstrated its effectiveness for use in downstream tasks in various domains, such as image \cite{frans2021clipdraw, gal2021stylegannada, Patashnik_2021_ICCV, rombach2022high}, 3D \cite{Michel_2022_CVPR, Sanghi_2022_CVPR, Wang_2022_CVPR}, and human motion \cite{motionclip, mdm, motiondiffuse}. Furthermore, MotionCLIP achieves zero-shot action classification performance close to 2s-AGCN \cite{agcn}, a dedicated action classifier, demonstrating the latent space's ability to discriminate between different action types \cite{motionclip}.

\begin{figure*}[!htbp]
    \centering
    \includegraphics[width=17.5cm]{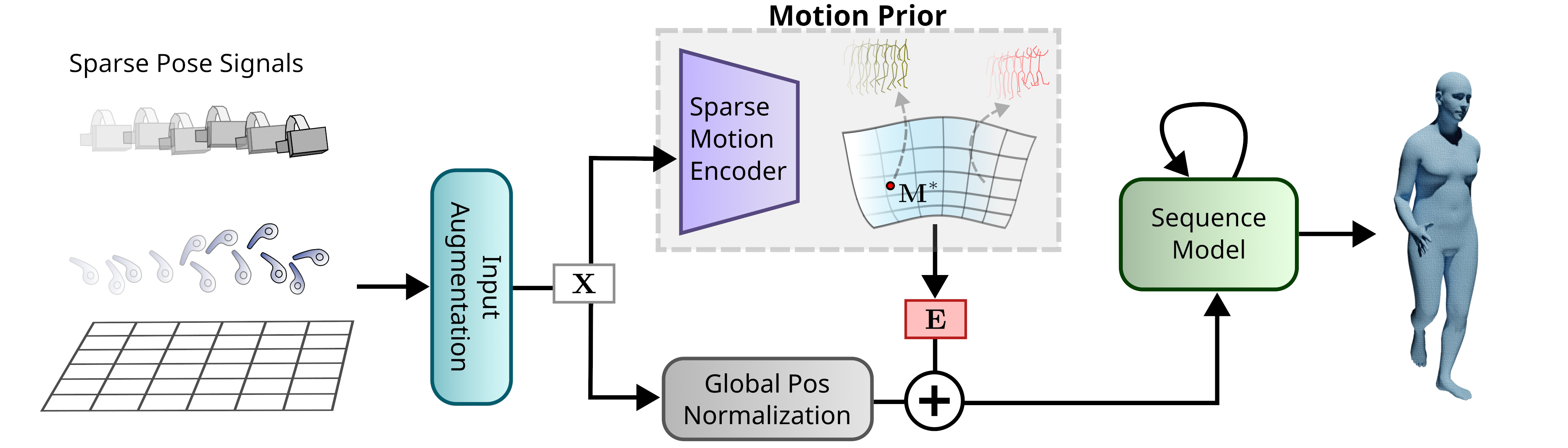}
    \caption{\textbf{Model Overview.} Sparse pose signals are fed to sparse motion encoder after augmentation to derive a motion embedding. Motion embeddings are combined with normalized augmented signals and input the sequence model for full-pose reconstruction.
    We tried different network architectures during development, and readers may refer to the supplementary material for details and experimental results.
    }
    \label{fig:model_overview}
\end{figure*}

\section{Methods}

\begin{figure}
    \centering
    \includegraphics[width=8.25cm, height=7.5cm]{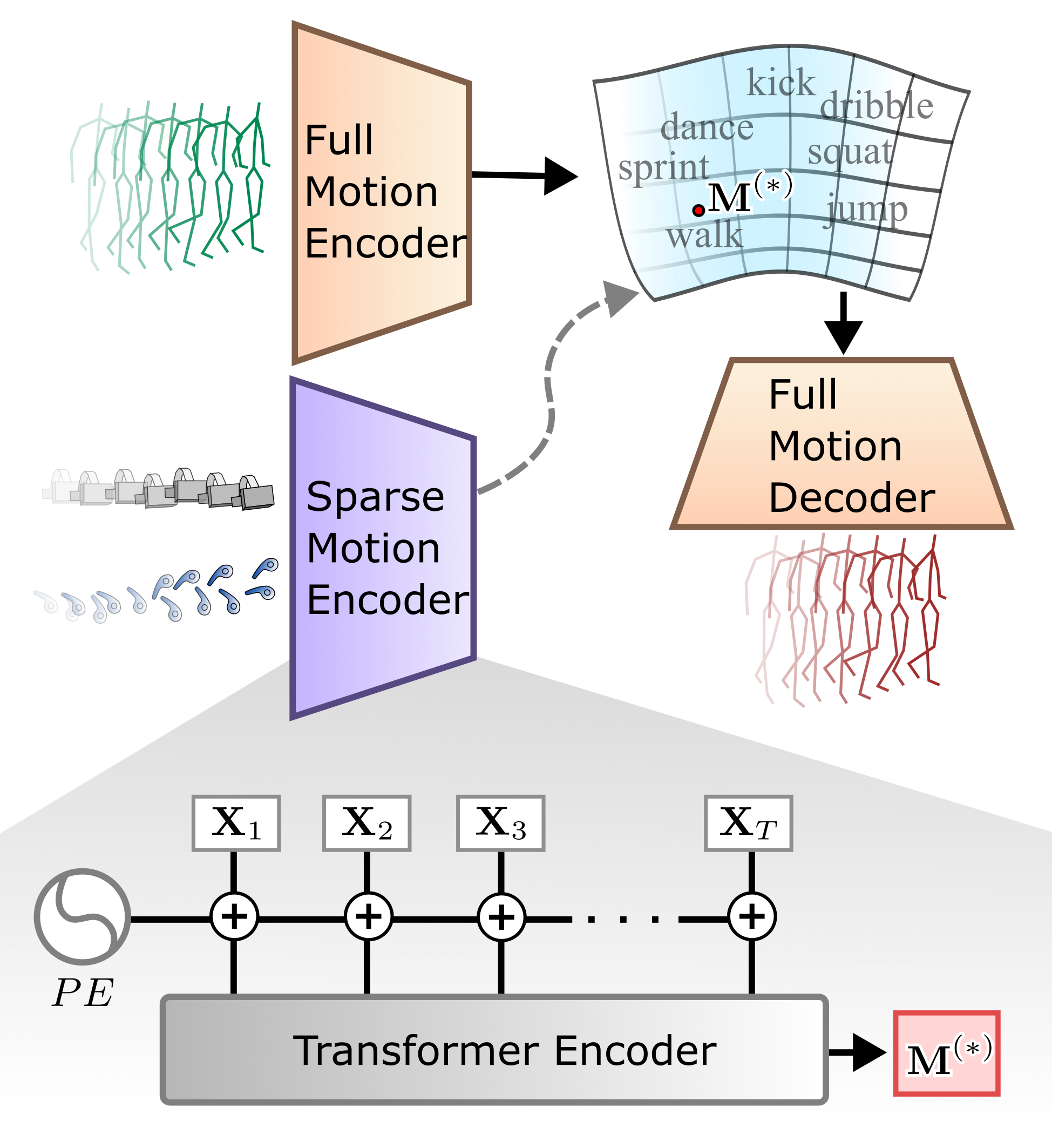}
    \caption{\textbf{Motion Prior.} After pretraining full motion prior, sparse motion encoder is trained to predict the same motion latent as the full motion encoder.}
    \label{fig:motion_prior}
\end{figure}

Our framework consists of the following components: \textit{full motion prior}, \textit{sparse motion encoder}, and \textit{sequence model}.
The full motion prior consists of full motion encoder and decoder (Figure \ref{fig:motion_prior}), trained on full-pose motions to learn the motion latent space. We train this component first, followed by training sparse motion encoder. The goal of the sparse motion encoder is to predict the \textit{motion latent} ($\mathbf{M}$) in the space learned by our full motion prior, utilizing only sparse pose signals.
Sparse motion encoder and sequence model are used directly for full-body pose estimation, the process visualized in Figure \ref{fig:model_overview}.
We feed the sparse pose signals to the sparse motion encoder after augmentation, from which we extract the \textit{motion embedding} ($\mathbf{E}$), which is a compressed representation of motion.
We then concatenate the motion embedding with the augmented sparse pose signals after normalization step. Finally, the concatenated sequence is input to sequence model for full-pose reconstruction.

\subsection{Input and Output Representations} \label{inout}
We represent the sparse pose signals at time $t$ as $\mathbf{x}_t$, defined as:
\begin{equation}
    \mathbf{x}_t = [ \mathbf{g}_t^{\mathbf{3}}, \mathbf{r}_t^{\mathbf{3}} ],
\end{equation}
where $\mathbf{g}_t^{\mathbf{3}}$ and $\mathbf{r}_t^{\mathbf{3}}$ respectively represent the global positions (3D) and the rotations in 6D form \cite{6drot} of head and hands, as would be provided by the MR device. From this data, we derive \textit{sparse motion signals} at time $t$ denoted $\mathbf{X}_t$, and a \textit{sparse motion sequence} of length $T$ at time $t$ denoted $\mathbf{X}_{t-T+1:t}$, each defined as:
\begin{equation}
    \mathbf{X}_t = [ \mathbf{g}_t^{\mathbf{3}}, \dot{\mathbf{g}}_t^{\mathbf{3}}, \mathbf{r}_t^{\mathbf{3}}, \dot{\mathbf{r}}_t^{\mathbf{3}} ] \in \mathbb{R}^{54},
\end{equation}
\begin{equation}
    \mathbf{X}_{t-T+1:t} = [ \mathbf{X}_{t-T+1}, \mathbf{X}_{t-T+2}, ..., \mathbf{X}_{t} ] \in \mathbb{R}^{54 \times T},
\end{equation}
where $\dot{\mathbf{g}}_t^{\mathbf{3}}$ and $\dot{\mathbf{r}}_t^{\mathbf{3}}$ respectively represent the velocities and angular velocities in 6D form derived from $\mathbf{g}_t^{\mathbf{3}}$ and $\mathbf{r}_t^{\mathbf{3}}$ as done in AvatarPoser \cite{avatarposer}.
This process is represented as ``Input Augmentation" in Figure \ref{fig:model_overview}.
The global position $\mathbf{g}_t^{\mathbf{3}}$ depends on an origin point decided by the MR device, which can be arbitrary. We counteract the randomness of input by ``normalizing" the global positions $\mathbf{g}_t^{\mathbf{3}}$ in the horizontal axes ($x$ and $z$ axes) as follows:
\begin{equation}
    \mathbf{g}_t^{\mathbf{3}} = [ \mathbf{g}_t^{head}, \mathbf{g}_t^{lhand}, \mathbf{g}_t^{rhand} ],
\end{equation}
\begin{equation}
    \bar{\mathbf{g}_t}.xz = \Bigl(\frac{\mathbf{g}_t^{head} + \mathbf{g}_t^{lhand} + \mathbf{g}_t^{rhand}}{3} \Bigr).xz,
\end{equation}
\begin{multline}
    \mathbf{n}_t^{j}.xz = (\mathbf{g}_t^{j} - \bar{\mathbf{g}_t}).xz, \quad
    \mathbf{n}_t^{j}.y = \mathbf{g}_t^{j}.y,
    \\ \forall j \in \{ head, lhand, rhand \},
\end{multline}
\begin{equation}
    \mathbf{n}_t^{\mathbf{3}} = [ \mathbf{n}_t^{head}, \mathbf{n}_t^{lhand}, \mathbf{n}_t^{rhand} ],
\end{equation}
where $\mathbf{n}_t^{\mathbf{3}}$ denote the normalized positions,
normalized positions of 3 joints having mean of 0 along $x$ and $z$ axes.
This process is represented as ``Global Pos Normalization" in Figure \ref{fig:model_overview}. We found empirically the information lost by using normalized global positions on the horizontal axes to be compensated for by the other inputs obtained via augmentation, and the normalization step to help the sequence model produce more stable motions. We did not observe the same benefit while training the sparse motion encoder, so we apply normalization only before the sequence model.

The output consists of relative rotations (6D form) for 22 SMPL \cite{smpl} joints, which can be used to recover the entire kinematic tree up to both wrists via forward kinematics (FK).

\subsection{Full Motion Prior Pretraining} \label{fmp}
Full motion prior denotes motion prior whose encoder and decoder are trained on full-pose sequences, i.e., full motions. We use MotionCLIP \cite{motionclip}, which is a full-body motion auto-encoder \cite{goodfellow2016deep} exploiting the powerful latent space of CLIP \cite{openaiclip}.
As visualized in Figure \ref{fig:motion_prior}, when the full 60-frame pose sequence is input to the encoder (based on the Transformer Encoder architecture \cite{transformer}), denoted \textit{full motion encoder}, the output is a latent vector lying in CLIP space, denoted \textit{motion latent} ($\mathbf{M}$). The decoder (based on the Transformer Decoder architecture \cite{transformer}), denoted \textit{full motion decoder}, aims to reconstruct the same full motion from the motion latent.
The loss $\mathcal{L}_{fm}$ used to train the full motion prior is formulated as follows:
\begin{equation}
    \mathcal{L}_{fm} = \mathcal{L}_{recon} + \lambda_{text} \mathcal{L}_{text} + \lambda_{image} \mathcal{L}_{image},
\end{equation}
\begin{equation}
    \mathcal{L}_{text} = 1 - \cos(\textit{CLIP}_{text}(t), \mathbf{M}),
\end{equation}
\begin{equation}
    \mathcal{L}_{image} = 1 - \cos(\textit{CLIP}_{image}(s), \mathbf{M}),
\end{equation}
where $\mathcal{L}_{recon}$ is the reconstruction loss of the full motion, and $\mathbf{M} \in \mathbb{R}^{512}$ represents the motion latent, which is the output of the full motion encoder.
$\mathcal{L}_{text}$ and $\mathcal{L}_{image}$ are the cosine distances from the motion latent $\mathbf{M}$ to its corresponding text projection $\textit{CLIP}_{text}(t)$, and the image projection $\textit{CLIP}_{image}(t)$, respectively.
Instead of learning a new latent space, as a variational auto-encoder \cite{vae} would do, CLIP space projection of the text labels $\textit{CLIP}_{text}(t)$ and that of the rendered images $\textit{CLIP}_{image}(s)$ corresponding respectively to each motion (both of which are part of dataset) are used to guide the motion latents to lie close together (with corresponding text and image projections) on the same space.
We train with the configuration named ``paper\_model" \cite{motionclipimpl}. This module is trained before all else.

\subsection{Estimating Motion Latent from Sparse Motion Sequence} \label{sparse_prior}
We train the sparse motion encoder to estimate the motion latent from sparse motion sequence $\mathbf{X}_{t-T+1:t}$. The architecture is adapted from full motion encoder's Transformer Encoder \cite{transformer} having the linear layer before the Transformer Encoder modified to accept $\mathbf{X}_{t-T+1:t} \in \mathbb{R}^{54 \times T}$. We use $T = 60$ which is the same motion length used by the full motion prior. We keep the full motion decoder and keep its weights frozen when we train the auto-encoder consisting of sparse motion encoder and full motion decoder (Figure \ref{fig:motion_prior}) with loss $\mathcal{L}_{sm}$ as follows:
\begin{equation}
    \mathcal{L}_{sm} = \lambda_{text}^* \mathcal{L}_{text}^* + \lambda_{image}^* \mathcal{L}_{image}^*,
\end{equation}
\begin{equation}
    \mathcal{L}_{text}^* = 1 - \cos(\textit{CLIP}_{text}(t), \mathbf{M}^*),
\end{equation}
\begin{equation}
    \mathcal{L}_{image}^* = 1 - \cos(\textit{CLIP}_{image}(s), \mathbf{M}^*),
\end{equation}
where $\mathbf{M}^* \in \mathbb{R}^{512}$ denotes the motion latent predicted by the sparse motion encoder. We set $\lambda_{text}^* = \lambda_{image}^* = 0.01$.
This module is trained after the full motion prior.

\begin{table*}
\small
\begin{center}
 \begin{tabular}{ |m{2.7cm}||m{2cm} m{2cm} m{1.5cm} m{1.5cm}|m{1.7cm} m{1.7cm}|  }
     \hline
     &\multicolumn{4}{|c|}{Per-Joint Errors} & \multicolumn{2}{|c|}{Motion-Related Statistics} \\
     \hline
     Method & MPJPE & Legs MPJPE & Global MPJPE & MPJVE & Motion \quad Distance $\downarrow$ & FID $\downarrow$ \\
     \hline 
     \hline
     Ours         & \textbf{7.25} & \textbf{9.34} & \textbf{7.38} & \textbf{25.42} & $\boldsymbol{5.12 \cdot 10^{-3}}$ & $\boldsymbol{6.03 \cdot 10^{-2}}$ \\
     AvatarPoser \cite{avatarposer}  & 7.71 & 10.25 & 7.79 & 29.71 & $5.38 \cdot 10^{-3}$ & $7.59 \cdot 10^{-2}$ \\
     AvatarPoser-60  & 7.74 & 10.39 & 7.82 & 29.85 & $5.47 \cdot 10^{-3}$ & $7.87 \cdot 10^{-2}$ \\
     VAE-HMD \cite{vaehmd} & 7.48 & \textbf{9.34} & 7.78 & 54.84 & $8.13 \cdot 10^{-3}$ & $9.24 \cdot 10^{-2}$ \\
     VAE-HMD-60 & 8.46 & 10.69 & 8.91 & 60.31 & $8.09 \cdot 10^{-3}$ & $9.40 \cdot 10^{-2}$ \\
     \hline
\end{tabular}
\end{center}
\caption{\textbf{Main Quantitative Results.}}
\label{table:main_results}
\end{table*}

\subsection{Sequence Model to Reconstruct Full-Pose from Sparse Motion Sequence and Motion Latent}
Sequence model takes as input the length-$S$ sparse motion sequence $\mathbf{X}_{t-S+1:t}$ and the corresponding sequence of \textit{motion embeddings} $\mathbf{E}_{t-S+1:t} = [ \mathbf{E}_{t-S+1}, \mathbf{E}_{t-S+2}, ..., \mathbf{E}_{t} ]$ as input. A motion embedding $\mathbf{E}_t \in \mathbb{R}^{64}$ is derived by passing the predicted motion latent $\mathbf{M}_t^* \in \mathbb{R}^{512}$ through a linear layer to retrieve a compressed 64-dimensional representation of the motion. Then, $\mathbf{X}_{t-S+1:t}$ and $\mathbf{E}_{t-S+1:t}$ are concatenated along the time axis, to be input to the sequence model.
3-Layer LSTM \cite{lstm} is our choice of sequence model, which outputs a single full-body pose at time $t$ given a sequence of inputs of length $S$ from time $t-S+1$ to $t$.
We represent the full-pose as the 6D relative rotation values of 22 joints of the SMPL model, from which we can recover the absolute rotations $\hat{\mathbf{r}}_t^\mathbf{22}$ and the body root-relative positions $\hat{\mathbf{p}}_t^\mathbf{22}$ of 22 joints via FK.

The loss $\mathcal{L}_{seq}$ is computed as the weighted sum of rotational loss $\mathcal{L}_{rot}$, positional loss $\mathcal{L}_{pos}$, velocity loss $\mathcal{L}_{vel}$ (ablation for $\mathcal{L}_{vel}$ in supplementary), and motion loss $\mathcal{L}_{mo}$. $\mathcal{L}_{rot}$, $\mathcal{L}_{pos}$, and $\mathcal{L}_{vel}$ encourage accurate full-pose reconstruction at every frame, and are computed as the L2 norm between the predicted and corresponding GT values, respectively.
The motion loss $\mathcal{L}_{mo}$ encourages the model to learn the correct motion given the consecutive 60-frame full-pose predictions $\hat{\mathbf{p}}_{t-60+1:t}^\mathbf{22}$, which are passed through a full motion encoder to obtain motion latent $\hat{\mathbf{M}}_t$. We also obtain the ground truth motion latent $\mathbf{M}_t$ with the corresponding ground truth motion $\mathbf{p}_{t-60+1:t}^\mathbf{22}$ via the same full motion encoder. While we use the pretrained full motion encoder in Section \ref{fmp} for this purpose, a different full motion encoder could substitute it. We then compute the cosine distance between the two motion latents to obtain motion loss:
\begin{equation} \label{modistloss}
    \mathcal{L}_{mo} = 1 - \cos(\hat{\mathbf{M}}_t, \mathbf{M}_t).
\end{equation}
Finally, the total loss of the sequence model $\mathcal{L}_{seq}$ is computed as follows:
\begin{equation}
    \mathcal{L} = \lambda_{rot} \mathcal{L}_{rot} + \lambda_{pos} \mathcal{L}_{pos} + \lambda_{vel} \mathcal{L}_{vel} + \lambda_{mo} \mathcal{L}_{mo}.
\end{equation}
We set the coefficients $\lambda_{rot} = \lambda_{pos} = \lambda_{vel} = 1.0, \lambda_{mo} = 0.1$.
Moreover, we found freezing the sparse motion encoder's weights to yield better results (Section \ref{ablation_studies}), and to allow preprocessing the motion latents in advance for faster training.

\section{Experimental Results}
\subsection{Data Preparation and Network Training}
We train and test all our models on the AMASS \cite{amass} dataset, a large-scale human motion dataset parametrized by the SMPL model \cite{smpl}. Since AMASS contains motion capture data with varying frame rates, we downsample each mocap data to be close to 30 FPS. From AMASS, we extract the head (joint index 15) and two wrist joints (joints indices 20 and 21) and derive their root-relative positions via FK, followed by adding translation to simulate global position signals given by MR devices. We also derive the absolute rotations of head and hands to simulate rotation signals.
Data are then ready to be processed by the procedure described in Section \ref{inout}. All of our model components and baselines share the same input and output format (n.b., while VAE-HMD \cite{vaehmd} was originally tested given pelvis-relative positions as input, we input global positions to reflect the signals from MR devices). We use the AMASS subset consisting of BMLrub \cite{amass_BMLrub}, EyesJapanDataset \cite{amass_Eyes_Japan}, TotalCapture \cite{amass_TotalCapture}, KIT \cite{amass_KIT}, ACCAD \cite{amass_KIT}, CMU \cite{amass_cmuWEB}, PosePrior \cite{amass_PosePrior}, TCDHands \cite{amass_TCD_hands}, EKUT for training and set aside HumanEva \cite{amass_HEva}, HDM05 \cite{amass_MPI_HDM05}, SFU \cite{amass_SFU}, MoSh \cite{amass_MoSh}, Transitions, SSM for evaluation. For training motion priors, we additionally use BABEL \cite{babel}, dataset containing per-frame action labels corresponding to a large portion of AMASS for the text labels, and images are rendered via MotionCLIP's official open-source implementation \cite{motionclipimpl}.

The full motion prior and the sparse motion encoder each takes 10 hours, and the sequence model 5 hours after preprocessing the motion latents via the trained sparse motion encoder (possible because sparse motion encoder is kept frozen during sequence model training) which takes about an hour, for a total 26 hours for full training on a single NVIDIA RTX 2080 Ti.
For the baseline models, we followed the setup described in the original papers \cite{vaehmd, avatarposer} as closely as possible. For AvatarPoser, we use the official open-source implementation \cite{avatarposerimpl}. For VAE-HMD \cite{vaehmd}, which has no open-source implementation available, we implemented their best performing model according to the original paper which contains a pretrained pose prior component. Note that we selected a much wider subset of AMASS for training and testing than the original works \cite{vaehmd, avatarposer}.
Refer to the supplementary material for more details about training and baseline implementations.

\subsection{Quantitative Evaluation} \label{quanteval}

We quantitatively evaluate our model using a diverse set of metrics against two baselines: AvatarPoser \cite{avatarposer} and VAE-HMD \cite{vaehmd}. The quantitative results are presented in Table \ref{table:main_results}.
Originally, AvatarPoser and VAE-HMD had window sizes of 40 frames and 16 frames, respectively. To ensure fairness, we additionally compare against AvatarPoser and VAE-HMD each adapted to have a 60-frame window size, which is the same as the window size that our motion prior sees. The adapted versions are labeled AvatarPoser-60 and VAE-HMD-60 in Table \ref{table:main_results} (further results and analyses in supplementary).

\begin{figure*}
    \centering
    \includegraphics[width=17.5cm, height=5.85cm]{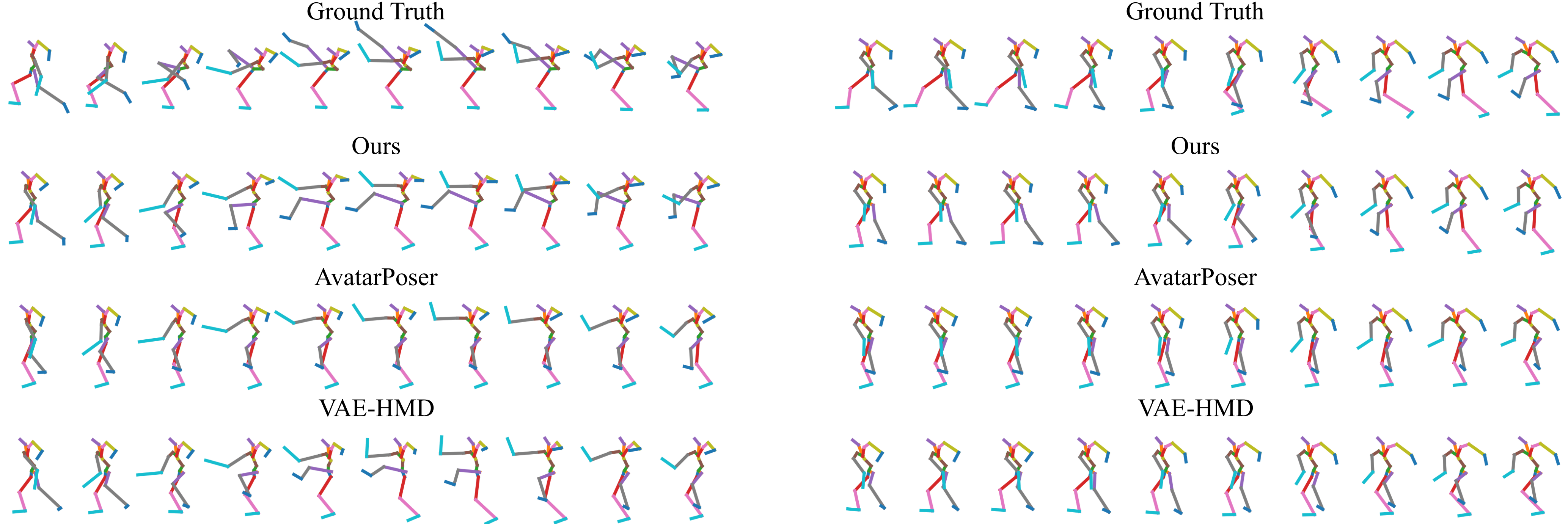}
    \caption{\textbf{Qualitative Results}. We show qualitative results on difficult motions with less common lower body movements. Left Column: Kicking Motion (Transitions/mazen\_c3d/kick\_push\_poses \cite{amass}). Right Column: Moonwalking (Transitions/mazen\_c3d/run\_stand\_poses \cite{amass}).}
    \label{fig:qual_eval_2anims}
\end{figure*}

\textbf{Per-Joint Errors}. We use four per-joint error metrics to evaluate our approach: MPJPE (mean per-joint position error [cm]), Legs MPJPE [cm], Global MPJPE [cm], and MPJVE (mean per-joint velocity error [cm/s]). Global MPJPE is computed by first mapping the predicted joints to global space, which involves combining GT head position (given by MR device) with the predicted joint rotations of the full body.
The results in Table \ref{table:main_results} demonstrate that our approach outperforms baselines on the majority of the metrics evaluated, rivalled only by VAE-HMD on Legs MPJPE. Since the sparse pose signals only contain direct information about the upper body, accurate reconstruction of legs is challenging especially when leg motions have low correlation with co-ocurring upper body motion.
Use of a motion prior generally results in lower error in leg motions, which can also be seen in Section \ref{ablation_studies} and in Figure \ref{fig:qual_eval_2anims} qualitatively. VAE-HMD contains a decoder component of a pretrained auto-encoder whose weights are frozen, and during pretraining, it receives full-body motions as input and learns a prior, which helps reconstruct some difficult motions, as evidenced by the legs MPJPE metric being as low as ours.
However, VAE-HMD suffers from relatively high velocity error (MPJVE) compared to other methods. 

\textbf{Motion-Related Errors}. Motion distance measures the difference between the overall motion (spanning 60 frames), between the GT motion and predicted motion calculated the same way as Equation \ref{modistloss}.
For this evaluation we use a different motion prior from one we optimized with, namely ``classes\_model" of MotionCLIP \cite{motionclip}).
Although classes\_model's motion latents also lie in CLIP space \cite{openaiclip}, different training parameters and dataset are used to train, resulting in different motion latents being predicted (Also, the text and images' CLIP projections do not align \cite{motionclip}, so different motion latent spaces are learned depending on training configuration).
FID (Fretchet Inception Distance \cite{fid}) measures the similarity between the distributions of ground truth motions and generated motions, the lower the more similar. 
While FID has been used widely in the context of generated human motions \cite{a2m, actor, motionvaes, nemf}, our work is the first to use it for full-body motions reconstructed from sparse signals. Our model achieves the lowest FID, followed by AvatarPoser variants, and VAE-HMD-60 with the highest FID.
These results are consistent with the findings from our user studies (Section \ref{user_study_results}) where we evaluate the quality of generated motions.

\begin{table}[!htbp]
\small
\begin{center}
 \begin{tabular}{ | m{2.5cm} || m{0.9cm}  m{0.9cm} | m{0.9cm}  m{0.9cm} | }
    \hline
      & \multicolumn{4}{m{5.0cm}|}{General Preference}    \\
      \hline
      & \multicolumn{2}{  m{2.5cm} | }{Random Set} & \multicolumn{2}{ m{2cm} | }{Hard Set} \\
     \hline
     Other Model & Ours & Other & Ours & Other \\
     \hline
     AvatarPoser & \textbf{58.8\%} & 41.2\% & \textbf{65.0\%} & 35.0\% \\
     VAE-HMD & \textbf{94.9\%} & 5.10\% & \textbf{95.8\%} & 4.20\% \\
     \hline
\end{tabular}
\end{center}
\caption{\textbf{Result of User Study I: General Preference.} (Better reconstruction of ground truth motion.) }
\label{table:survey1_results_pref}
\end{table}

\begin{table}[!htbp]
\small
\begin{center}
 \begin{tabular}{ | m{1.7cm} || m{0.65cm} m{0.65cm} m{0.7cm} | m{0.65cm} m{0.65cm} m{0.7cm} | }
    \hline
      & \multicolumn{6}{m{4.0cm}|}{Motion Matching}    \\
      \hline
      & \multicolumn{3}{  m{2cm} | }{Random Set} & \multicolumn{3}{ m{2cm} | }{Hard Set} \\
     \hline
     Other Model & Ours & Other & Neut & Ours & Other & Neut \\
     \hline
     AvatarPoser & \textbf{43.9\%} & 32.8\% & 23.3\% & \textbf{52.6\%} & 28.8\% & 18.6\% \\
     VAE-HMD     & \textbf{76.9\%} & 10.3\%  & 12.8\% & \textbf{79.8\%} & 12.7\% & 7.50\% \\
     \hline
\end{tabular}
\end{center}
\caption{\textbf{Result of User Study I: Motion Matching.} (Which generated motion matches ground truth motion better?)}
\label{table:survey1_results_mm}
\end{table}

\begin{figure}[!htbp]
    \centering
    \includegraphics[width=1.0\linewidth]{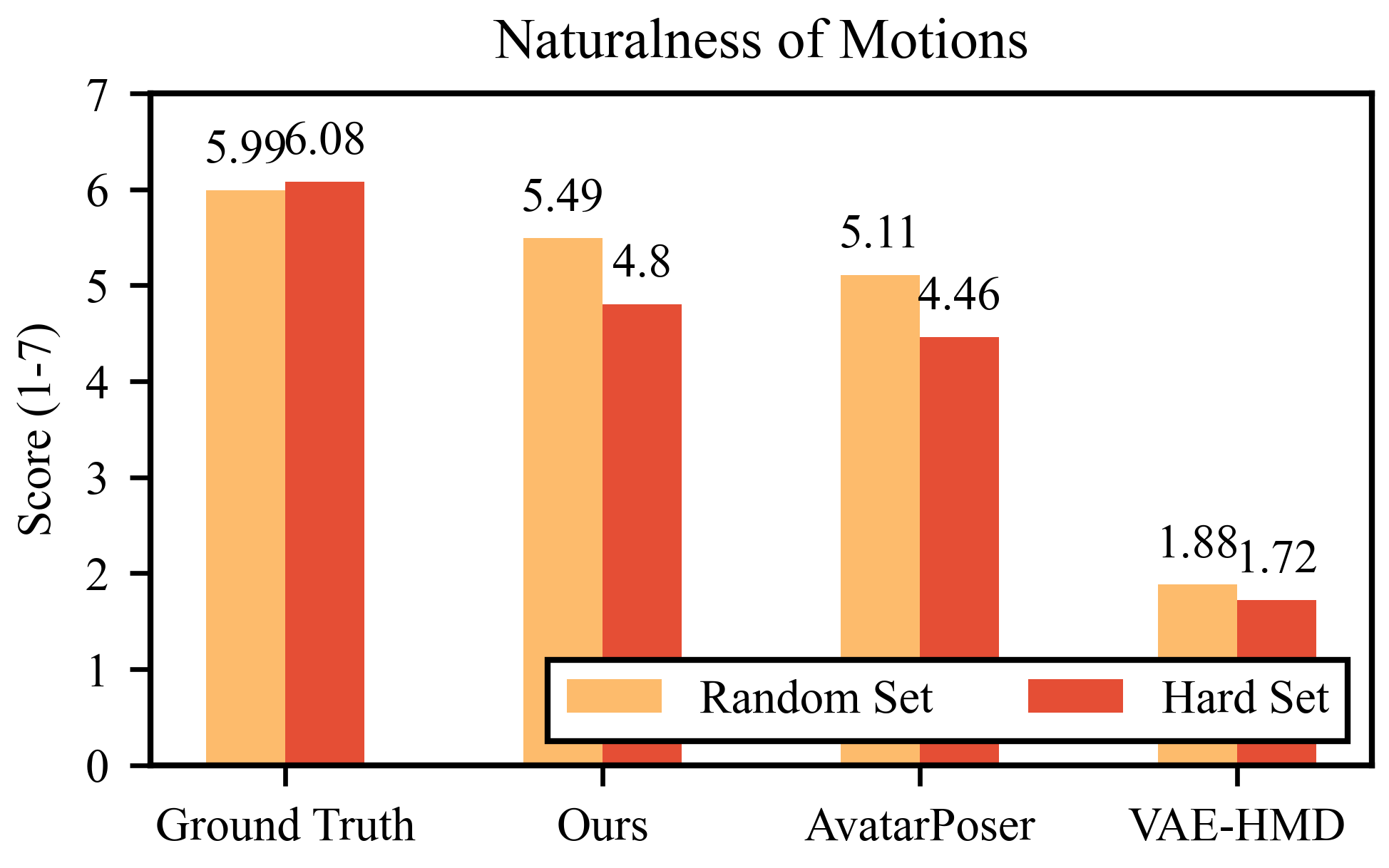}
    \caption{\textbf{Result of User Study II: Naturalness of Motions.} Scores range from 1 (worst) to 7 (best).}
    \label{fig:survey2_results}
\end{figure}

\subsection{User Study} \label{user_study_results}
We define two sets of motion segments for the user studies: Random Set, consisting of 60 3-second segments randomly sampled from the entire test dataset, and Hard Set, which includes 60 3-second segments. For Hard Set, we sampled 30 motion segments on which one of our baselines (neither AvatarPoser nor VAE-HMD, but the ``No Motion Prior" model explained in Section \ref{ablation_studies}), trained without motion prior, had the highest MPJPE and Legs MPJPE each. We conducted user studies I and II, where for each user study, we (disjointly) sampled 30 segments, from Random Set and Hard Set respectively. We conducted the user studies on 30 participants. Full-body animations of ground truth motions, predictions from our model, AvatarPoser \cite{avatarposer}, and VAE-HMD \cite{vaehmd} for each segment were rendered using SMPL Blender addon \cite{smplblenderplugin}. 

\textbf{User Study I}. 
To compare our model with one of the baselines side-by-side, we placed ground truth animation on top and juxtaposed our model's prediction and one of AvatarPoser and VAE-HMD's animations at the bottom (latter's order randomized) for the same underlying motion. We informed participants that the two animations at the bottom are different reconstructions of the top animation from partial information, and we asked to choose (1) a better reconstruction among the two (preference), and (2) one whose motion matched the ground truth motion better, with ``neutral" option added for the latter.

The results for (1) can be found in Table \ref{table:survey1_results_pref}. The table shows participants' preference towards our model's predictions for both sets, with more pronounced results with Hard Set. This demonstrates our model's ability to better reconstruct motions that the baseline, which does not make use of a motion prior, struggles with. Moreover, participants clearly preferred our model's predictions over VAE-HMD's, noting that VAE-HMD's animations looked unnatural primarily due to jitter. This is consistent with the high velocity error (MPJVE) measured in Section \ref{quanteval}.
The results of (2) can be found in Table \ref{table:survey1_results_mm}, where we evaluated the motion matching capability of our model in addition to the quantitative motion distance metric in Section \ref{quanteval}. The results show similar trends as (1).

\textbf{User Study II}. 
We played all four animations with the same underlying motion simultaneously, ordered randomly, asking participants to rate the naturalness of each motion on a scale of 1 to 7, 7 being the highest (participants were allowed to replay animations as they desired). From the mean scores plotted in Figure \ref{fig:survey2_results}, we can observe that users found GT motions the most natural, followed by generations from our model's predictions, then AvatarPoser and VAE-HMD, in that order. While the scores for predicted motions for Hard Set fall behind those for Random Set, participants found the motions generated via our model more natural than other models' generations for both sets.


\section{Ablation Studies} \label{ablation_studies}

\begin{table*}
\small
\begin{center}
 \begin{tabular}{ |m{5.55cm}||m{1.5cm} m{1.5cm} m{1.5cm} m{1.5cm}|m{1.7cm} m{1.7cm}|  }
     \hline
     &\multicolumn{4}{|c}{Per-Joint Errors} & \multicolumn{2}{|c|}{Motion-Related Statistics} \\
     \hline
      Method & MPJPE & Legs MPJPE & Global MPJPE & MPJVE & Motion \quad Distance $\downarrow$ & FID $\downarrow$ \\
     \hline 
     \hline
     Ours         & \textbf{7.25} & 9.34 & \textbf{7.38} & \textbf{25.42} & $\boldsymbol{5.12 \cdot 10^{-3}}$ & $\boldsymbol{6.03 \cdot 10^{-2}}$ \\
     \hline
     No Motion Prior  & 7.37 & 9.67 & 7.62 & 26.22 & $5.52 \cdot 10^{-3}$ & $7.54 \cdot 10^{-2}$ \\
     No Motion Distance Loss & 7.32 & 9.43 & 7.45 & 25.71 & $5.31 \cdot 10^{-3}$ & $7.08 \cdot 10^{-2}$ \\
     With Finetuned Motion Prior  & 7.39 & 9.77 & 7.67 & 26.10 & $5.28 \cdot 10^{-3}$  & $6.24 \cdot 10^{-2}$ \\
     With a Different Motion Prior & 7.29 & \textbf{9.27} & 7.41 & 26.15 & $5.15 \cdot 10^{-3}$ & $6.20 \cdot 10^{-2}$ \\
     \hline
\end{tabular}
\end{center}
\caption{\textbf{Ablation Studies: Main Quantitative Results}}
\label{table:ablation_quant_results}
\end{table*}

\begin{table}
\small
\begin{center}
 \begin{tabular}{ | m{3.6cm} || m{3.6cm} | }
    \hline
    Improved Action Type & Degraded Action Type \\
    \hline \hline
     \multicolumn{2}{ | m{7cm} |}{Legs MPJPE Improvement/Degradation} \\
     \hline
     \textbf{knee movement} & place something  \\
     \textbf{cartwheel}     & grasp object     \\
     \textbf{crouch}        & \textbf{poses}   \\
     \textbf{squat}         & stretch          \\
     bend                   & face direction   \\
     \hline
     \multicolumn{2}{ | m{7cm} |}{MPJPE Improvement/Degradation} \\
     \hline
     \textbf{cartwheel}     & face direction  \\
     \textbf{shuffle}       & place something  \\
     \textbf{knee movement} & \textbf{lean}     \\
     throw                  & take/pick something up  \\
     \textbf{touch ground}  & shout      \\
     \hline
\end{tabular}
\end{center}
\caption{\textbf{Ablation Studies: Action Types \cite{babel} Improved by Using Motion Prior.} Actions containing high amount of leg motions are in bold.
Our model's top improvements lie in actions containing much leg motions.
}
\label{table:improved_actiontypes}
\end{table}

We conduct ablation studies by removing or modifying different subcomponents, the main quantitative results shown in Table \ref{table:ablation_quant_results}.

We first assess the role of the motion prior component in our architecture by completely removing it from our model, leaving only the input processing and sequence model components (see Figure \ref{fig:model_overview} for reference). The results in ``No Motion Prior" row in Table \ref{table:ablation_quant_results} show degradation in values for all quantitative metrics. (This model was used to curate the Hard Set for the user study, as explained in Section \ref{user_study_results}.)
Additionally, we group all motion segments in the test dataset by action types defined in BABEL \cite{babel}, and sort them by improvement of Legs MPJPE and MPJPE respectively. We present the top 5 and bottom 5 action types in Table \ref{table:improved_actiontypes}, left column showing top 5 and right column showing bottom 5. Top 5 improved action types for both metrics consist of actions involving a high amount of leg motions \footnote{A list of BABEL \cite{babel} action subtypes corresponding to each action type can be found in \cite{babelactionsubtypes}.}, showing that the motion prior contributes to better reconstruction of leg motions given only upper body signals.
We can also observe degradation from not using the motion distance loss (``No Motion Distance Loss") and from unfreezing the motion prior (``With Finetuned Motion Prior”). 

We experimented with various task-generic motion priors during development and settled on MotionCLIP as it gave the best overall performance. The final row of Table \ref{table:ablation_quant_results} shows the result of using a Transformer VAE-based \cite{transformer, vae} motion prior whose architecture is based on ACTOR \cite{actor}, from which we removed action conditioning part to have an unconditional motion VAE \cite{vae}.

\section{Conclusions and Limitations}
We present a method of utilizing motion prior to effectively reconstruct full-body motion from impoverished signals of pose. Our method recovers intended full-body motions that look natural, with improved lower body over baselines.
However, our work only considers a single body type, and we wish to allow people of diverse body shapes to utillize our system effectively in a future work.
Moreover, we sometimes observe footsliding artifacts from generated motions, and we wish to measure their severity and eliminate them.

\clearpage
{\small
\bibliographystyle{ieee_fullname}
\bibliography{bibliography}
}

\end{document}